\documentclass{ifacconf}

\usepackage[caption=false,font=footnotesize]{subfig}

\usepackage{graphicx}      
\usepackage{natbib}        
\usepackage{amsmath}

\usepackage{booktabs}   
\usepackage{siunitx}    
\usepackage{caption}    
\usepackage{float}
\usepackage{stfloats}

\usepackage[nolist,nohyperlinks]{acronym}
\begin{acronym}
\acro{DVL}{Doppler Velocity Logger}
\acro{FFT}{Fast Fourier Transform}
\acro{GNSS}{Global Navigation Satellite System}
\acro{IMR}{Inspection, Maintenance and Repair}
\acro{LSQ}{Least Squares}
\acro{MC-Lab}{Marine Cybernetic Laboratory}
\acro{ROI}{Region of Interest}
\acro{ROV}{Remotely Operated Vehicle}
\acro{SLAM}{Simultaneous Localization and Mapping}
\acro{ROS}{Robot Operating System}
\acro{USBL}{Ultra Short Baseline}
\acro{UUV}{Unmanned Underwater Vehicle}
\end{acronym}

\begin{document}
\begin{frontmatter}

\title{A Vision-based Control Framework for Real-time Autonomous UUV Operations\thanksref{footnoteinfo}} 

\thanks[footnoteinfo]{Corresponding author: {\tt\small marco.job@ntnu.no}}

\author[First,Second]{E. T. Frøland}
\author[First]{M. Job}
\author[First]{MDE. Deowan}
\author[First]{E. Kelasidi}

\address[First]{Dept. of Mechanical and Industrial Engineering, NTNU, Norway}
\address[Second]{Oceaneering AS, Norway}

\begin{abstract}
This paper presents a fully integrated vision-based framework for real-time and robust localization, autonomous navigation, and mapping for unmanned underwater vehicles (UUVs) in dynamic, visually challenging environments.
The proposed pipeline enables both net-relative and global localization while generating continuous 3D maps of the surroundings in real-time.
The framework was validated on synthetic datasets with ground truth and tested onboard an UUV during autonomous net-relative navigation experiments.
Results demonstrate real-time performance and enhanced robustness, supporting vision-driven autonomous navigation and enabling the field deployment of marine robots for critical inspection and mapping tasks in complex underwater environments.
\end{abstract}


\end{frontmatter}
\section{Introduction}
Robotic systems are increasingly explored for \ac{IMR} in sea-based aquaculture, where operations are labor-intensive and hazardous~\citep{evjemo2026aquaculture}.
While \acp{ROV} and semi-automated systems exist, they still rely heavily on manual control and expert supervision.
Fully autonomous \acp{UUV} promise continuous net inspection, hole detection, and environmental monitoring~\citep{kelasidi2022autonomous, madshaven2022hole}, but achieving this requires robust perception, localization, and mapping in dynamic and visually degraded underwater environments.
In subsea operations, \ac{GNSS} is unavailable and external infrastructure is typically absent~\citep{evjemo2026aquaculture}.
These environments are particularly challenging due to low visibility, light scattering and moving objects, making automation difficult and demanding.
Addressing these challenges is the focus of this paper.

Traditionally, UUVs rely on acoustic sensors such as \ac{USBL} and \ac{DVL} for underwater localization and navigation~\citep{kelasidi2022robotics,rundtop2016experimental}.
However, in aquaculture settings these systems struggle due to signal interference from fish, weak reflections from flexible net-pens, and continuous net-pen deformations~\citep{rundtop2016experimental,amundsen2022autonomous}.
Furthermore, their dependence on dedicated equipment installations and overall additional cost makes them less appealing in sectors where cost-effectiveness is essential~\citep{evjemo2026aquaculture}.
While low-cost laser-based alternatives have demonstrated accuracy comparable to DVL systems, they still face limitations similar to acoustic-based solutions~\citep{bjerkeng2023absolute}.
As an alternative, various vision-based approaches have been explored for underwater navigation, including stereo vision~\citep{skaldebo2024approaches}, FFT-based image processing~\citep{schellewald2021vision}, and learning-based methods such as TRUDepth~\citep{ebner2024metrically}, which offer cost-efficient depth estimation, but remain sensitive to turbidity and low-light conditions.
\ac{SLAM} and mapping techniques play a central role in autonomous operation~\citep{singh2025deepvl,cardaillac2023application}, yet state-of-the-art systems like SVIn2 and TURTLMap~\citep{wang2023realtime, song2024turtlmap} still struggle in low-texture, repetitive and unstructured environments.
Approaching the problem in a decoupled manner, net-pen specific localization methods combined with volumetric mapping frameworks offer a robust alternative for specific underwater environments.
A framework for robust localization and mapping proposed by~\cite{job2025leveraging} integrated an FFT-based depth perception method, TRUDepth, and wavemap~\citep{reijgwart2023wavemap} and demonstrated strong potential, but did not yet support real-time, fully onboard operation.

This paper extends the framework proposed in~\cite{job2025leveraging} by integrating the developed methods into a complete pipeline for real-time, fully onboard autonomous navigation and mapping for UUVs in complex net-pen environments.
The proposed framework is based on the \ac{ROS} ecosystem, and key components include an FFT-based method for sparse depth estimation, an optimized TRUDepth node for dense depth estimation, a relative pose node for net-relative localization, and a global pose node that fuses \ac{DVL} and the vision-based estimates to compute the UUV's position within the net-pen coordinate system.
The pipeline is integrated on an embedded system onboard an actual UUV for autonomous navigation.
It is evaluated using synthetic data with ground truth and further assessed for real-time performance prior to experimental validation.
Simulation and real-world experiments demonstrate the framework's efficacy and robustness, enabling future deployments in underwater structure monitoring and inspection tasks.
To the authors' best knowledge, this is the first work to demonstrate fully onboard and autonomous localization, navigation, and mapping in net-pen aquaculture environments, highlighting the unique contribution and practical relevance of the proposed approach.

\section{Proposed Framework}
\vspace{-2mm}
\label{sec:proposed_framework}

\begin{figure*}[ht!]
    \centering
    \includegraphics[width=1\textwidth]{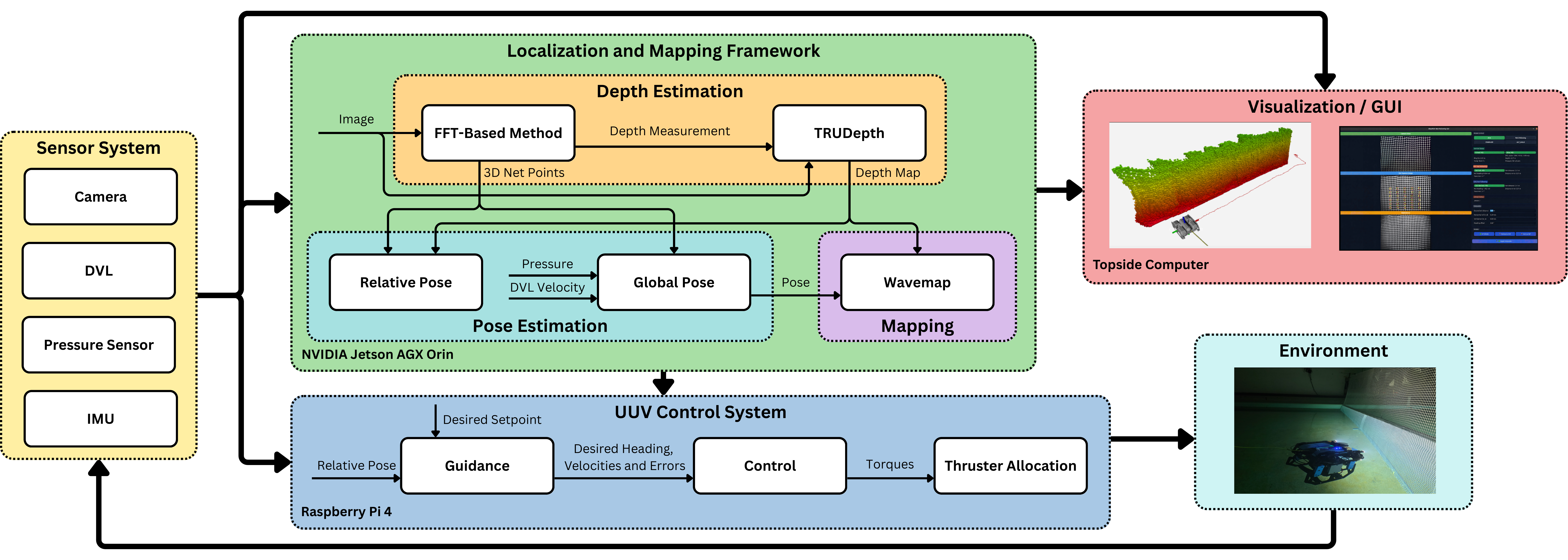}
    \caption{Overview of the framework architecture.}
    \vspace{-2mm}
    \label{fig:framework_overview}
\end{figure*}

This section presents the complete framework for real-time monocular-vision-enhanced localization, autonomous navigation, and mapping.
The framework is composed of interconnected modules designed for operations with UUVs, with the overall objective of estimating the vehicle's relative and global pose while reconstructing a detailed 3D map of the surrounding net-pen.
Each module is implemented as an independent \ac{ROS} node, enabling modular development, parallel execution, and integration with the UUV's onboard sensor systems.
Fig.~\ref{fig:framework_overview} provides an overview of the framework architecture, which consists of five main nodes: the FFT-based sparse depth estimation node, the TRUDepth dense depth estimation node, the relative pose node, the global pose node, and wavemap.
The FFT-based method extracts 3D net points and depth measurements from camera images, which are then used by the TRUDepth node to generate dense pixel-wise depth maps.
The net-relative pose and global pose modules estimate the UUV's orientation and position by leveraging visual and acoustic sensor-based measurements.
Finally, the wavemap module fuses dense depth maps with global pose estimates to incrementally construct a volumetric 3D map of the net-pen.
Together, these components form a continuous real-time and onboard processing pipeline, where data from the UUV's sensors flow through the interconnected \ac{ROS} nodes to produce synchronized localization and mapping results.
In the following, the main components and adaptations compared to~\cite{job2025leveraging} will be discussed; we refer to the original works for details on the theory.
\subsection{FFT-based Sparse Net 3D Points}
\begin{figure}[ht]
\vspace{-2mm}
      \centering
      \subfloat[Detected net squares \label{DetectedNetSquares}]
      {
            \includegraphics[scale=0.13]{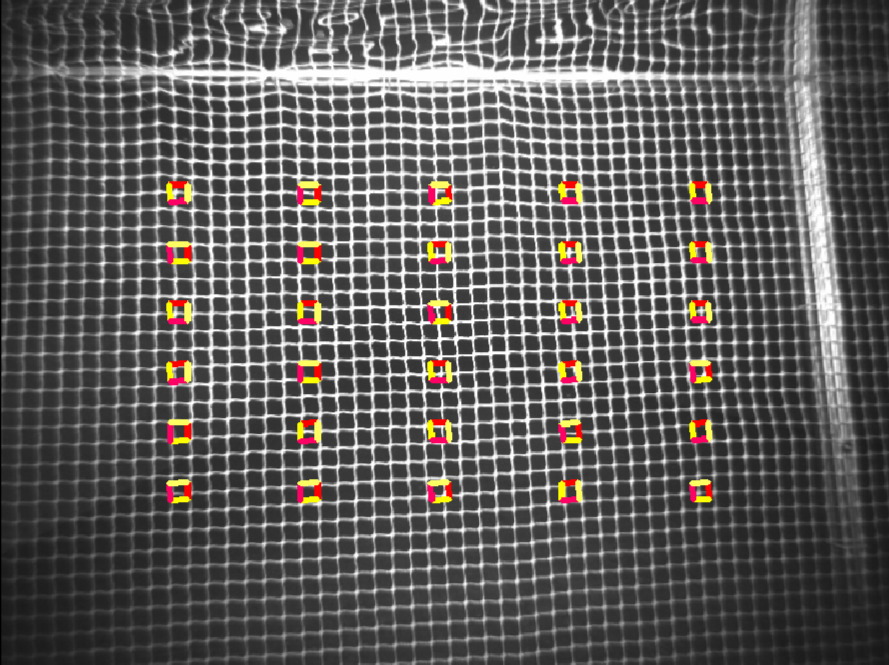}
      }
      \subfloat[Depth map \label{DepthMap}]
      {
            \includegraphics[scale=0.13]{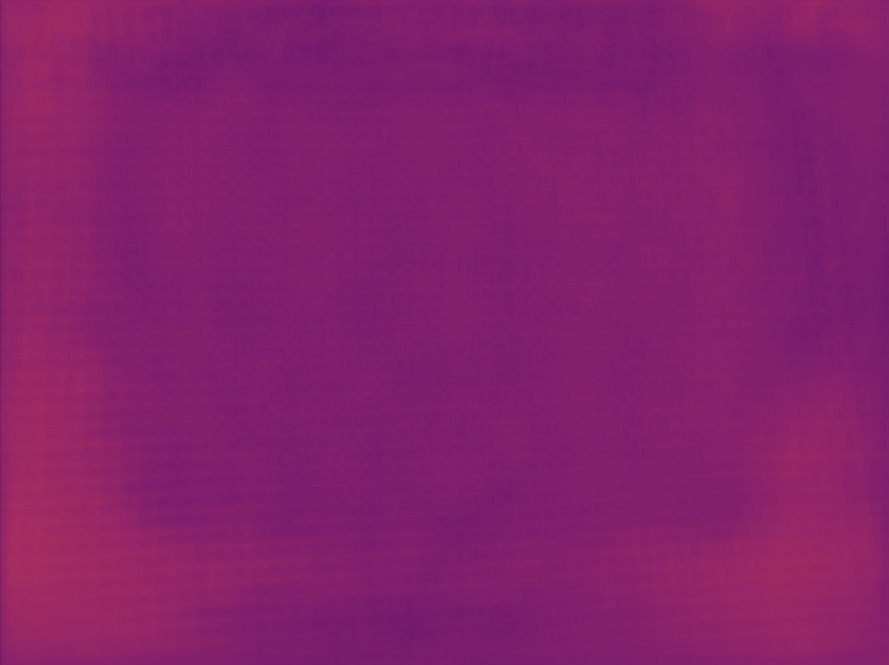}
      }
      \caption{Outputs from the FFT-based (a) and TRUDepth (b) methods at 1\,m distance.}
\end{figure}

\vspace{-2mm}
The FFT-based method identifies the periodic pattern in the net grid and, together with the known grid size, computes 3D points of the net surface in the camera frame~\citep{schellewald2021vision}.
Each image is divided into squares and overlapping \acp{ROI}.
The size and number of \acp{ROI} are configurable parameters that trade off detection accuracy and computational load.
Together with known camera parameters and net grid size, a net-relative pose per \ac{ROI} in camera frame is estimated (Fig.~\ref{DetectedNetSquares}).
This is used to reconstruct a sparse set of 3D points on the net surface, with size proportional to the number of ROIs.
The 3D points are also used to provide sparse depth measurements in image space, which are triples of pixel coordinates and corresponding depth values $(p_{x}, p_{y}, d)$.
Both outputs serve as inputs for the subsequent TRUDepth, relative pose, and global pose modules.
In this integrated pipeline, the method is implemented as a \ac{ROS} node that subscribes to the real-time camera image stream, computes the 3D net points and depth measurements, and publishes both outputs.
The outputs inherit the camera timestamps to ensure synchronized processing in downstream modules.
\vspace{-3mm}
\subsection{TRUDepth}
\vspace{-2mm}
The TRUDepth method generates metric dense depth maps using depth completion on RGB images with FFT-based depth measurements~\citep{job2025leveraging}.
These measurements are converted into two dense inputs, a nearest-neighbor depth map $S_1(x, y)$, and a probability map $S_2(x, y)$.
Implemented as a \ac{ROS} node, it subscribes to camera images and point-cloud measurements, synchronizing them by timestamp to ensure that each RGB frame is paired with its corresponding measurements.
The network then performs inference to produce a pixel-wise metric depth (Fig.~\ref{DepthMap}), published for use in downstream modules.

In addition to accuracy, real-time operation is a key requirement for the proposed framework in this paper.
The main performance bottleneck in the TRUDepth node of previous works was the generation of the dense input maps $S_1(x, y)$, and $S_2(x, y)$.
The proposed optimization process replaced the original loop-based computation of distance maps with a fully vectorized approach.
This allowed the TRUDepth node to achieve real-time performance on embedded GPU platforms.

\vspace{-1mm}
\subsection{Relative Pose}
\vspace{-2mm}
Estimating the \ac{UUV}'s net-relative pose is crucial for navigation and control~\citep{job2025leveraging}, and this framework employs two vision-based approaches as alternatives to \ac{DVL}-based acoustic fitting~\citep{amundsen2022autonomous}.
The net-relative distance $d_{net}$ is defined as the forward distance from the UUV to the net along the camera z-axis. The net-relative heading $\psi_{net}$ corresponds to the relative angle between the camera z-axis and the estimated normal of the net surface, projected onto the camera xz-plane.
The first method uses the 3D net points obtained from the FFT-based pipeline to fit a plane via \ac{LSQ} for estimating heading $\psi_{net, FFT}$, and a paraboloidal surface to infer the net-relative distance $d_{net, FFT}$.
The second method leverages the dense depth maps from TRUDepth, computing distance $d_{net, TRUDepth}$ as the mean depth in a central image region and deriving the heading $\psi_{net, TRUDepth}$ by projecting a subsampled set of depth pixels into 3D using the camera intrinsics before fitting a plane in the same manner as FFT-based 3D net points.
Both approaches are integrated into a dedicated relative pose node, where each method is processed by separate callbacks but shares common underlying functions to maintain consistency.
The node publishes net-relative distance, pitch, and yaw on dedicated topics.

\subsection{Global Pose}
\vspace{-2mm}
The UUV's global pose relative to a fixed global reference frame centered at the net-pen surface is obtained under the common assumption that the net-pen is a perfect cylinder and the UUV maintains zero roll and pitch~\citep{job2025leveraging}.
The radial distance $r_g$ and net-relative yaw $\psi_r$ are derived by projecting FFT-based 3D net points onto the global $xz$-plane and fitting a circle using the known net-pen diameter.
The global angular coordinate $\rho_g$ is obtained by integrating horizontal velocities from the DVL, while depth $z_g$ comes from the depth sensor.
The full global position is expressed as $(r_g,\rho_g,z_g)$ with orientation $(0,0,\psi_g)$, where $\psi_g = \rho_g + \psi_r$, providing a complete global pose estimate within the net-pen.
The global pose node manages this estimation process by receiving input from the FFT-based node, the DVL, and the pressure sensor (Fig.~\ref{fig:framework_overview}).
It subscribes to the 3D net points expressed in camera frame, DVL velocity measurements, and depth readings, and synchronizes the incoming data streams to ensure temporal alignment.
The node estimates and publishes the \ac{UUV}'s global position and orientation.
\subsection{Wavemap}
\vspace{-2mm}
Wavemap~\citep{reijgwart2023wavemap} is a hierarchical, multi-resolution mapping framework with specific emphasis on efficiency and real-time performance, making it well-suited for UUV applications.
The framework constructs a volumetric map of the net-pen in real-time by combining dense depth maps from the TRUDepth node with inputs from the global pose node, aligning each measurement with its corresponding pose via timestamps to ensure spatial and temporal consistency.
The mapping runs continuously as the UUV moves, producing an incrementally updated and progressively refined occupancy grid.
Wavemap is configured for the specific dense depth sensing modality, setting minimum and maximum depth limits, camera model and intrinsics.
\subsection{UUV Control System}
\vspace{-2mm}
To enable autonomous net-relative path following, the control law proposed by~\cite{amundsen2022autonomous} was integrated into the pipeline, with the relative pose node providing the required inputs.
For low-level control, a PID velocity controller was implemented.
Since the proposed vision-based method does not provide velocity estimates, two strategies were used:
a) using the DVL as a velocity source while using the vision-based estimates for pose, or
b) providing the controller with constant PWM commands to the thrusters.
This allows a fully vision-based control framework, though this results in open-loop vehicle velocity control leading to higher sensitivity to disturbances.
As shown in Fig.~\ref{fig:framework_overview}, dedicated nodes were also implemented to visualize the net-relative and global pose estimations, depth map images, and the 3D volumetric occupancy grid in real time.
\vspace{-2mm}
\section{Results}
\vspace{-2mm}
This section first discusses the synthetic net-pen dataset created to evaluate the proposed framework with accurate ground truth.
Then the real-world experimental setup and the results are presented. Fig.~\ref{fig:framework_overview} shows the full system architecture deployed during the experiments.
\begin{figure}[ht]
    \centering
    \includegraphics[width=1.0\linewidth]{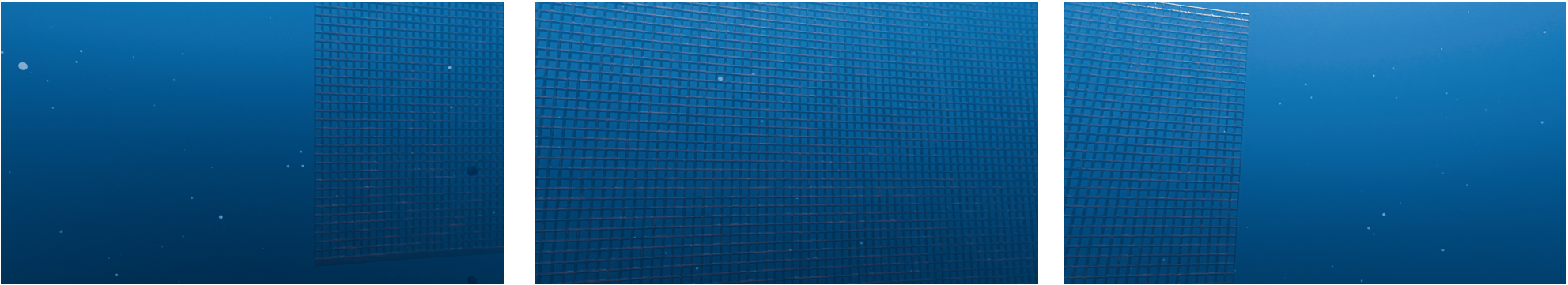}
    \caption{Three sample frames from the synthetic net-pen dataset, with realistic visual underwater conditions.}\vspace{-3mm}
    \label{SyntheticData}
\end{figure}
\begin{figure*}[h!]
\centering
\subfloat[Net-relative distance \label{NetRelativeDistance}]{
        \includegraphics[width=0.32\linewidth]{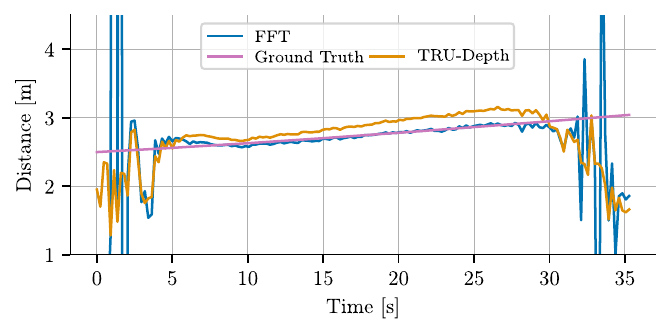}
      }
\subfloat[Net-relative heading \label{NetRelativeHeading}]{
        \includegraphics[width=0.32\linewidth]{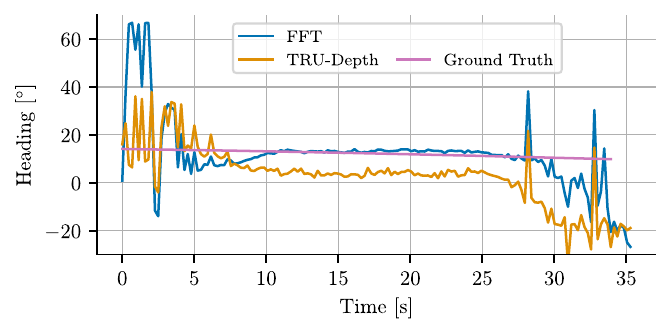}
      }
\subfloat[Global pose \label{GlobalPose}]{
        \includegraphics[width=0.32\linewidth]{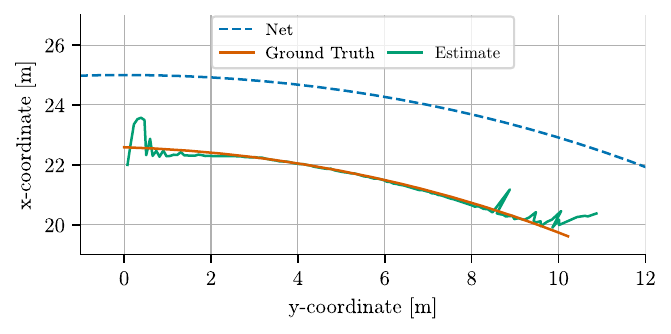}
      }\vspace{-2mm}
\caption{Net-relative distance, heading and global pose estimates from the synthetic dataset.}
\label{SyntheticDataResults}
\vspace{-2mm}
\end{figure*}
\subsection{Synthetic Net-Pen Dataset}
\vspace{-2mm}
In response to the challenge of obtaining accurate ground truth in the underwater domain, this paper also creates a synthetic dataset, which allows a more rigorous assessment of the framework's reliability and accuracy.
The synthetic photorealistic net-pen is created in Blender and the \ac{UUV} motion trajectory is defined to vary in orientation and distance to the net-pen.
The scene shows a realistic underwater environment with turbidity, suspended particles, and depth-dependent light attenuation (see Fig.~\ref{SyntheticData}).
Relative poses, global poses and velocities were extracted for each frame using the ground truth data of the simulation.

Fig.~\ref{SyntheticDataResults} shows the corresponding results for net-relative pose and global pose estimations.
Overall, both relative and global pose estimation methods exhibit larger errors at the beginning and end of the sequences, where the net is mostly outside the image frame.
In these intervals, only a small fraction of net points is detected, making plane fitting less reliable.
As the net occupies the majority of the image, the estimates are more precise and closely follow the ground truth, indicating that performance is primarily limited by net visibility rather than noise or rendering conditions.
The FFT-based method consistently aligns more closely with ground truth for both distance and heading, as it relies on prior knowledge of the mesh grid size and extracts points directly from the net surface.
In contrast, TRUDepth reconstructs dense depth maps including background and non-net structures, making plane fitting inherently less reliable.
In summary, both methods perform comparably, with the FFT-based approach showing superior consistency and accuracy.

\subsection{Experimental Setup}
\vspace{-2mm}
Experiments were performed at the \ac{MC-Lab} in Trondheim, Norway.
The facility features an indoor basin with dimensions of $40\,\text{m} \times 6.4\,\text{m} \times 1.5\,\text{m}$.
To approximate a fish farm net-pen, a net structure with dimensions of $4.9\,\text{m} \times 1.5\,\text{m}$ and net grid size of $30\,\text{mm} \times 30\,\text{mm}$ was built and installed.
The built net structure is flat, and the approach requires prior knowledge of the net-pen radius for global pose estimation.
In these experiments, we set the knowledge of the net-pen radius to $100\,\text{m}$, and assume that the curvature is negligible over the net width.
A BlueROV2 in heavy configuration has been utilized as the UUV platform for the experiments.
It is equipped with an IMU, a pressure sensor, a Nortek Nucleus 1000 \ac{DVL} and an Alphasense multi-camera system.
The \ac{DVL} was mounted forward-facing, and the multi-camera system contains five monochrome cameras, yet for this work only the front left camera was used.
\begin{figure}[h]
\centering
{
    \includegraphics[height=4.5cm]{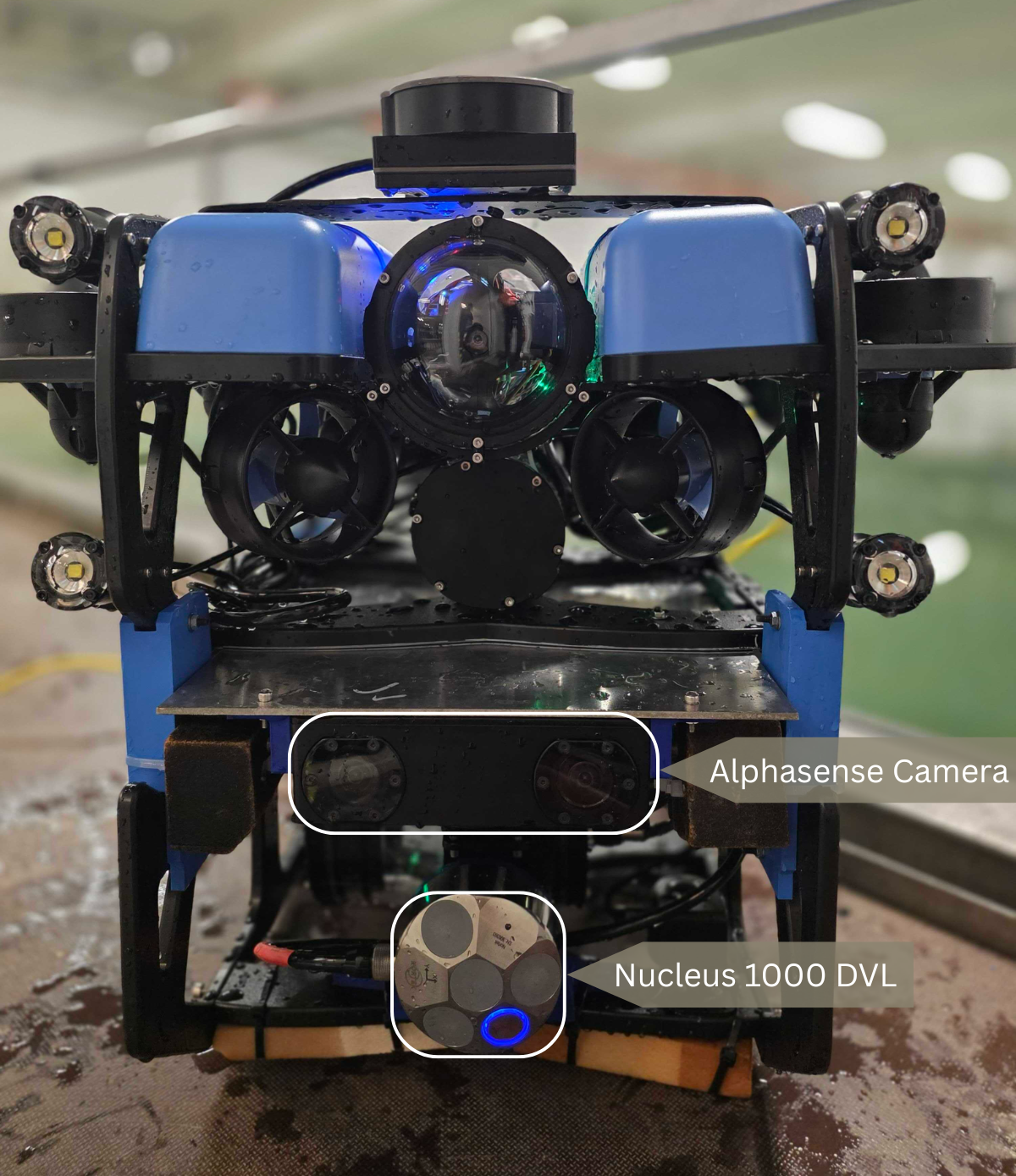}
}
\caption{BlueROV2 with integrated sensors.}
\label{UUVIntegratedSystemExperiments}
\end{figure}

Fig.~\ref{fig:framework_overview} shows the overview of the full UUV system.
All sensor drivers and all guidance, navigation, and control algorithms were running on a Raspberry Pi 4.
The camera driver and the vision-based pipeline presented in Section \ref{sec:proposed_framework} were running on an NVIDIA Jetson AGX Orin for real-time processing on the embedded GPU.
Vehicle communication for feedback during the experiments was handled using the Pymavlink MAVLink message-processing library, while interaction between the various onboard systems was managed through ROS.
The joystick interface for manual control and visualizations were the only components running on the topside computer.

Prior to the autonomous experiments, the intrinsic, extrinsic and distortion parameters of the Alphasense camera were calibrated inside the pool.
In addition, the FFT-based method parameters were tuned to balance detection accuracy in a desired range, given the camera resolution and field of view.
Table \ref{tab:perception_params} summarizes the main parameters used during the experiments.
Finally, due to differences in hardware and in the geometric interpretation of the net-relative distances, modifications were made to the DVL plane-approximation procedure in~\cite{amundsen2022autonomous}:
All net-pen distances in our experiments refer to the forward-facing distance to the net instead of the Euclidean distance to the closest point on the net plane.
For the DVL, the distance is defined as the distance derived from the plane that is spanned by the two vectors connecting the three DVL beam endpoints.

\renewcommand{\thetable}{\arabic{table}}
\captionsetup[table]{labelformat=simple, labelsep=colon, name=Tab.}
\begin{table}
    \centering
    \caption{Parameters related to FFT-based method.}
    \label{tab:perception_params}
    \begin{tabular}{lc}
        \toprule
        Parameter & Value \\
        \midrule
        Alphasense resolution            & $720\times540$\,\unit{px}\\
        Alphasense focal length          & $462$\,\unit{px}\\
        Alphasense frame rate            & $10$\,\unit{Hz}\\
        \midrule
        (ROI rows, ROI cols)             & (5, 6)\\
        ROI square size                  & $300$\,\unit{px}\\
        \bottomrule
    \end{tabular}
\end{table}
\begin{table}
\centering
\caption{PID gains used for each controlled degree of freedom.}
\begin{tabular}{lccc}
\hline
DOF & $K_p$ & $K_i$ & $K_d$ \\
\hline
$x$ (surge position)     & 3.0   & 0.0    & 0.3   \\
$y$ (sway position)      & 30.0  & 0.001  & 0.0   \\
$\psi$ (yaw)             & 0.075 & 0.007  & 0.008 \\
$u$ (surge velocity)     & 10.0  & 1.0    & 0.0   \\
\hline
\end{tabular}
\label{tab:pid_gains}
\end{table}

\begin{table}
\centering
\caption{Details on processing time and frequency on an NVIDIA Jetson AGX Orin.}
\begin{tabular}{lcc}
\hline
Node & Mean processing time & Frequency\\
\hline
Alphasense camera        & N/A & 10\,Hz\\
Nortek DVL               & N/A & 4\,Hz\\
FFT                      & 47.8\,ms & 10\,Hz\\
TRUDepth                 & 66.8\,ms & 10\,Hz\\
Relative Pose (FFT)      & 13.2\,ms& 10\,Hz\\
Relative Pose (TRUDepth) & 44.6\,ms & 10\,Hz\\
Global Pose              & 8.7\,ms & 4\,Hz\\
wavemap                  & N/A     & 0.5\,Hz\\
\hline
\end{tabular}
\label{tab:time}
\end{table}

\subsection{Experimental Results}
\vspace{-2mm}
\begin{figure*}[ht!]
\centering
\subfloat[Net-relative distance \label{NetRelativeDistance05}]{
        \includegraphics[scale=0.34]{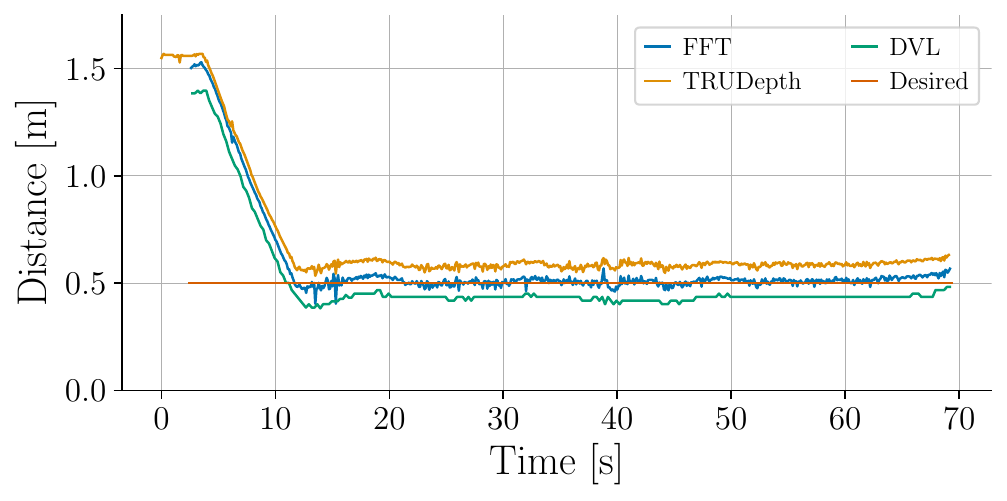}
      }
\subfloat[Net-relative heading \label{NetRelativeHeading05}]{
        \includegraphics[scale=0.34]{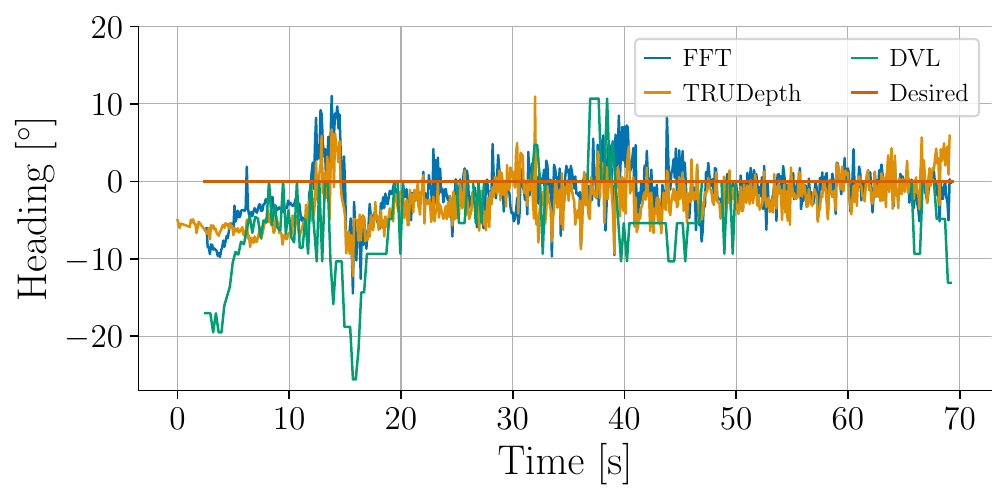}
      }
\subfloat[Global pose \label{GlobalPose05}]{
        \includegraphics[scale=0.34]{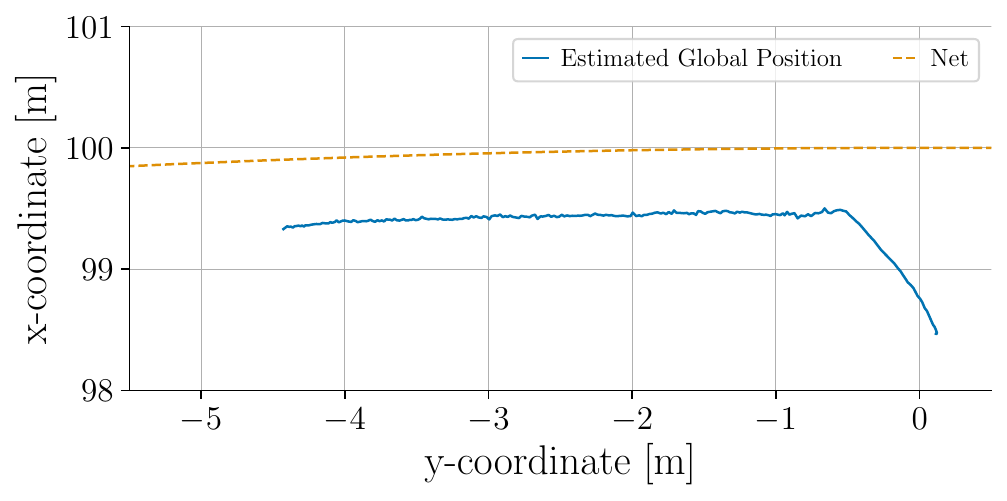}
      }
\caption{Autonomous net-relative navigation of the UUV keeping a desired distance of 0.5\,m from the net-pen.}
\label{NetRelativeNavigation05}
\end{figure*}
\begin{figure*}[ht!]
\centering
\subfloat[Net-relative distance \label{NetRelativeDistancelawnmower}]{
        \includegraphics[scale=0.34]{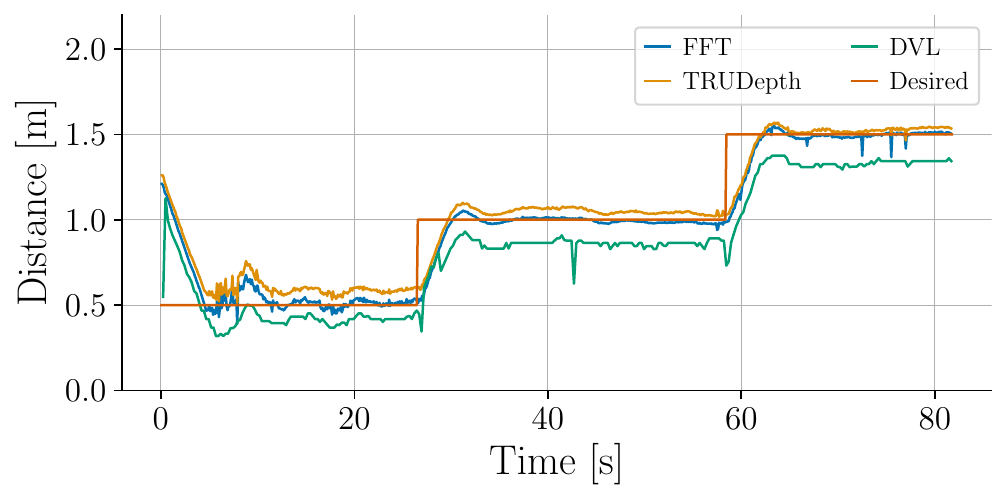}
      }
\subfloat[Net-relative heading \label{NetRelativeHeadinglawnmower}]{
        \includegraphics[scale=0.34]{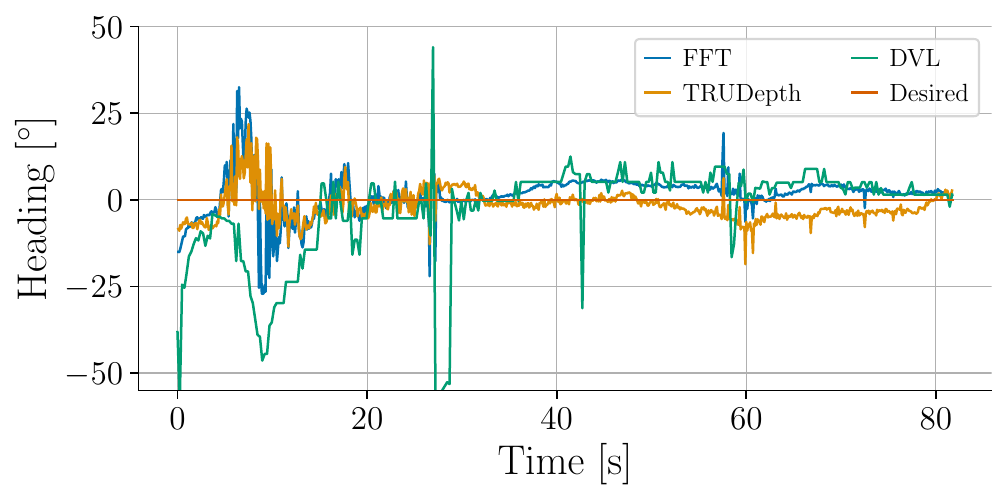}
      }
\subfloat[Global pose \label{GlobalPoselawnmower}]{
        \includegraphics[scale=0.34]{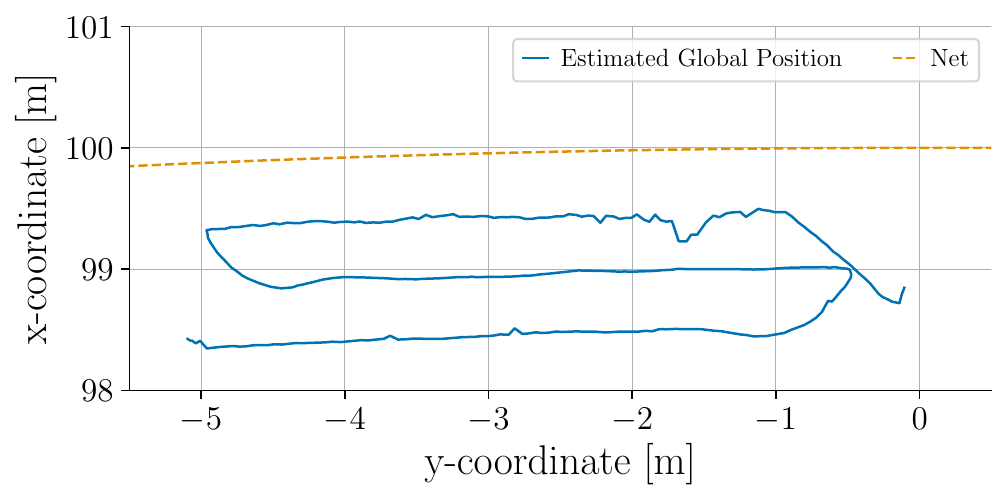}
      }
\caption{Autonomous net-relative navigation of the UUV when performing a "lawnmower" pattern, i.e., 0.5\,m, 1\,m, and 1.5\,m.}
\label{NetRelativeNavigationlawnmower}
\end{figure*}

Several experiments were performed to evaluate the performance and robustness of the proposed framework.
The UUV was commanded to keep varying desired net-relative distances, sideways speeds, and motion patterns under different onboard and external lighting conditions.
All experiments were performed using the Ardusub auto depth and roll-pitch stabilization mode~\citep{haugalokken2024lowcost}.
The gains used for the net-relative guidance control system~\citep{amundsen2022autonomous} during the trials are summarized in Tab. \ref{tab:pid_gains}.

The camera image acquisition rate determined the frame rate of the FFT-based method for 3D net grid estimation, TRUDepth and relative pose nodes.
The global pose node operated at the DVL update rate, while the wavemap node updated asynchronously based on the overall inputs.
The acquisition rate per sensor and mean processing times per node are summarized in Tab. \ref{tab:time}.

Fig.~\ref{NetRelativeNavigation05} and Fig.~\ref{NetRelativeNavigationlawnmower} show the response of the UUV when commanded to follow the desired distance (i.e., $0.5~\text{m}$) and perform a lawnmower pattern with varying reference distances (e.g., $0.5~\text{m}$, $1~\text{m}$, and $1.5~\text{m}$) from the net-pen with zero commanded net-relative heading.
The FFT-based method estimated the actual distance and orientation with high precision, and the UUV followed the desired reference signals through the guidance and control scheme in both cases.
The results are compared to the estimated distances from the TRUDepth method and the \ac{DVL} plane approximations.
Fig.~\ref{NetRelativeDistance05} and Fig.~\ref{NetRelativeDistancelawnmower} clearly illustrate that the TRUDepth slightly overestimates the distance, which is in line with the ground truth results obtained from the synthetic data between the FFT-based and TRUDepth methods (Fig.~\ref{NetRelativeDistance}).
In addition, the \ac{DVL} underestimated the distance from the net-pen, and further investigation should be conducted in the future to compare the \ac{DVL} and FFT-based results.
The volumetric maps were generated using wavemap (see the top-right of Fig.~\ref{fig:framework_overview}), based on depth images from TRUDepth and the corresponding 3D pose estimates (Fig.~\ref{GlobalPose05} and Fig.~\ref{GlobalPoselawnmower}).

\begin{figure}[!h]
\centering
\subfloat[FFT detection \label{low_light_fft_image}]{
        \includegraphics[scale=0.10]{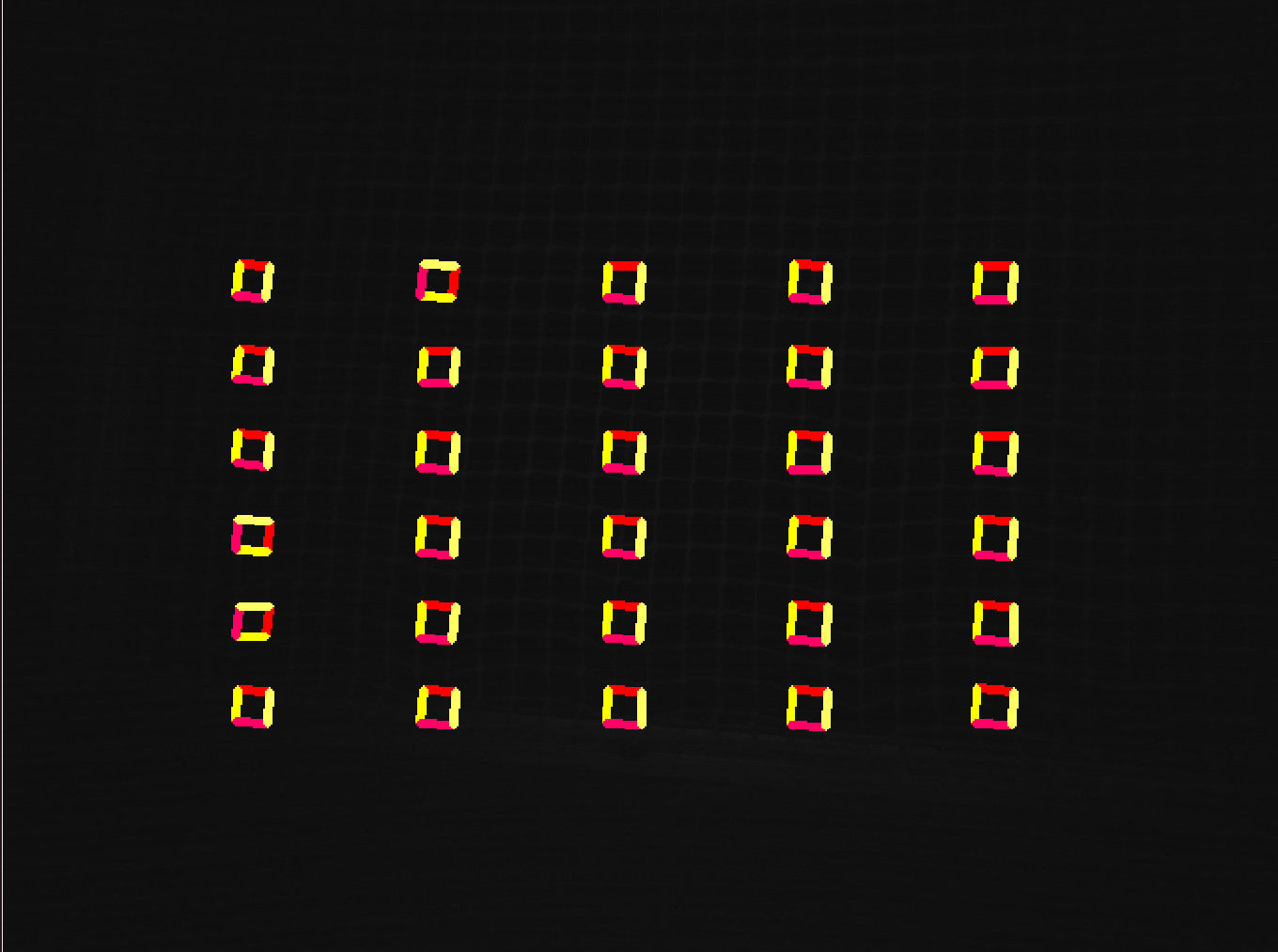}
      }\\\vspace{-2mm}
\subfloat[ Net-relative distance \label{low_light_relative_distance}]{
        \includegraphics[scale=0.45]{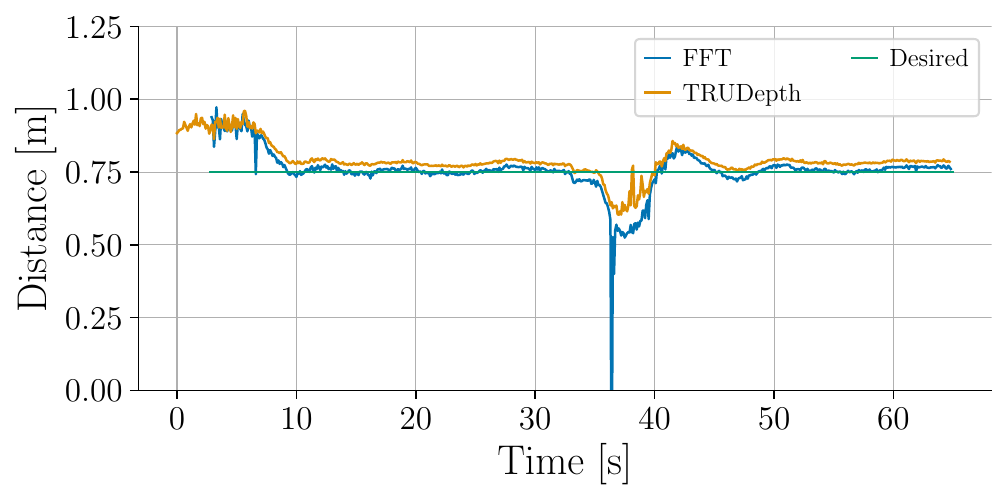}
      }\\\vspace{-2mm}
\subfloat[ Net-relative heading \label{low_light_relative_heading}]{
        \includegraphics[scale=0.45]{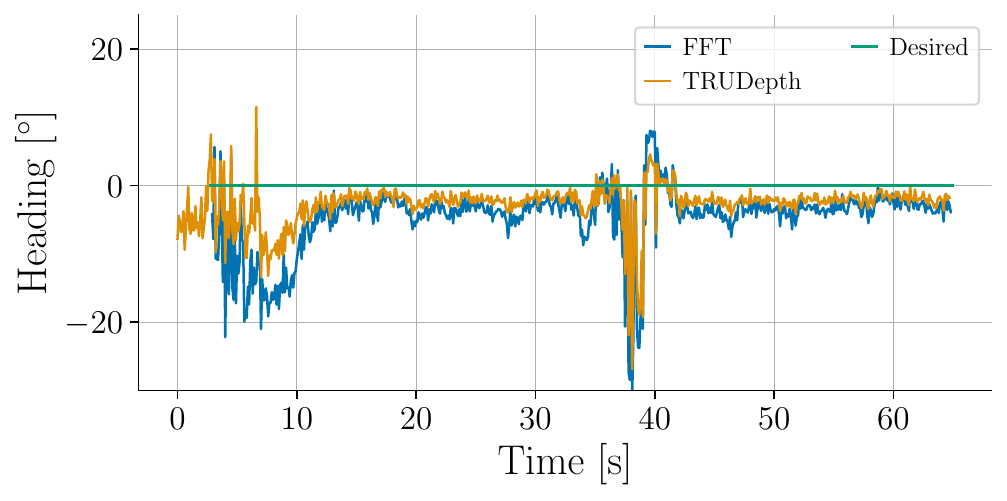}
      }
\caption{Experimental results during operations in low-light conditions while performing net-following at 0.75\,m.}
\end{figure}

A further noteworthy result involves testing the framework under extreme low-light conditions, where all external pool lights were turned off, and the UUV onboard lights were set to 25\% of the maximum brightness.
Fig.~\ref{low_light_fft_image} shows a sample image from the Alphasense camera during this low-light run, where the FFT-based method was still able to detect the net pattern, and estimate the net-relative distance and heading, shown in Fig.~\ref{low_light_relative_distance} and \ref{low_light_relative_heading}.
The spike in the distance estimate between 35 and 40\,s stems from a spike in the \ac{DVL} velocity measurement.
Despite this, the system demonstrated robustness and re-stabilized rapidly.

\section{Conclusions and Future Work}
\vspace{-2mm}
The integrated vision-based control framework presented in this paper combines monocular depth prediction with sparse FFT-based depth measurements to estimate net-relative pose and generate dense depth maps from vision alone.
These estimates are fused with acoustic measurements to obtain the global pose of the UUV, while the predicted depth images are incorporated into the wavemap framework for real-time 3D reconstruction.
The resulting ROS-based architecture forms a cohesive, modular, and real-time-capable system for onboard processing of continuous sensor streams during UUV operations, making it suitable for real-world field deployment.
The modular design allows new sensors and algorithms to be integrated; the FFT node could be replaced by alternative image-based depth measurements or by 3D point clouds supplied by an imaging sonar.
Experimental results demonstrate accurate real-time UUV localization, autonomous net-relative path following, and high-quality volumetric reconstructions suitable for inspection tasks.
Overall, the framework provides a robust and extensible foundation for real-time underwater localization, autonomous navigation and mapping, meeting the operational needs of aquaculture inspection while supporting future research and development in autonomous underwater systems.

\begin{ack}
\vspace{-2mm}
This work was supported by the Research Council of Norway (ResiFarm: NO-327292, CHANGE: N313737).
\end{ack}
\vspace{-2mm}
\bibliography{bibliography}
\end{document}